\documentclass{svproc}
\usepackage{url}

\usepackage{graphicx}

\begin{document}
\mainmatter              
\title{Multimodal and Crossmodal AI \\ for Smart Data Analysis}
\titlerunning{Multimodal and Crossmodal AI for Smart Data Analysis}  
%
\author{Minh-Son Dao}
\authorrunning{Minh-Son Dao} 
%
\tocauthor{Minh-Son Dao}
\institute{Big Data Integration Center\\National Institute of Information and Communications Technology \\
\email{dao@nict.go.jp},\\ WWW home page:
\texttt{https://bdirc.nict.go.jp/}
}

\maketitle              

\begin{abstract}
Recently, the multimodal and crossmodal AI techniques have attracted the attention of communities. The former aims to collect disjointed and heterogeneous data to compensate for complementary information to enhance robust prediction. The latter targets to utilize one modality to predict another modality by discovering the common attention sharing between them. Although both approaches share the same target: generate smart data from collected raw data, the former demands more modalities while the latter aims to decrease the variety of modalities. This paper first discusses the role of multimodal and crossmodal AI in smart data analysis in general. Then, we introduce the multimodal and crossmodal AI framework (MMCRAI) to balance the abovementioned approaches and make it easy to scale into different domains. This framework is integrated into xDataPF \footnote{the cross-data platform https://www.xdata.nict.jp/}. We also introduce and discuss various applications built on this framework and xDataPF.
\keywords{multimodal, crossmodal, AI, data analytics}
\end{abstract}
\section{Introduction}
\label{INTRO}

We daily struggle with processing large amounts of (un)intentionally-collected raw data (e.g., statistics, numbers, texts, images, audio) to get insights from our world. Nevertheless, smart data is a type of data we want to have instead of dealing with raw data containing redundant, even useless information. Smart data results from raw data’s analysis and interpretation, making it possible to draw value from it effectively. Hence, we need intelligent layers to embed in data collectors and storage to produce such smart data for further downstream applications. The process that turns a set of raw data into smart data could be considered smart data analytics. We can see many algorithms, products, and techniques using the prefix "smart" to express that they have smart data in their products, such as smart IoT, smart dashcams, and smart clouds. 

Human beings have cognition of the surrounding world by sensing from different perspectives (e.g., see, hear, smell, feel, taste). Hence, devices made by human beings tend to record/capture data of the surrounding world in the same way human beings do. Each type of data recorded/generated by a particular device/method represents how something happens or is experienced, and that representative can be concerned as a modality. A research problem or dataset that includes multiple such modalities is considered multimodal, and AI techniques that deal with multimodal are called multimodal AI. 

The advantage of multimodal is that we can have joint representative space that can compensate for the lack of information on each disjoint modality and strengthen the robust prediction of high-correlation modalities. Hence, we can build models that process and correlate data from multiple modalities.

Many surveys have been done to understand the use of multimodal AI for smart data analysis. In \cite{multimodal_Baltrusaitis2019}, the authors list out challenges of multimodal machine learning (e.g., representation, translation, alignment, fusion, co-learning), data types (e.g., texts, videos, images, audios) and applications (e.g., Speech recognition and synthesis, Event detection, Emotion and affect, Media description, Multimedia retrieval). In \cite{multimodal_bayoudh2022}, the authors emphasize a particular domain - computer vision, and introduce advances, trends, applications, and datasets of multimodal AI. In this survey, the authors discuss the general architecture of multimodal deep-learning, where a particular feature extraction first precedes each modality to create a modality representation. Then, these representations are fused into one joint representative space and project this space into one unique similarity measure. Several deep learning models are concerned in this survey, including ANN, CNN, RCNN, LSTM, etc. 

In \cite{modal_Vukotic2016}, the authors mention crossmodal learning for dealing with the issue when there is a need for mapping from one modality to another and back, as well as representing them in joint representation space. This direction is similar to human beings' learning process - composers a global perspective from multiple distinct senses and resources. For example, text-image matching, text-video crossmodal retrieval, emotion recognition, and image-captioning are the most popular crossmodal applications where people can use one modality to query another one \cite{crossmodal_wang2016}\cite{crossmodal_Ji202}\cite{crossmodal_Khare2021}. The main difference between multimodal and crossmodal learning is that crossmodal requires sharing characteristics of different modalities to compensate for the lack of information towards enabling the ability to use data of one modality to retrieve/query/predict data of another modality. Unfortunately, this research direction is far from the expectation and has a big gap among research teams and domains \cite{crossmodal_zhang2022}. 

In light of the abovementioned discussions, we are conducting research and development to build a multimodal and crossmodal AI framework for smart data analysis. The framework aims to provide additional intelligent layers to data analysis progress that can flexibly change from using only multimodal AI, crossmodal AI, or hybrid multi-crossmodal AI for analyzing data. We also introduce several instances of this framework designed for a particular domain, such as air pollution forecast, congestion prediction, and traffic incident querying. 

\section{Multimodal and Crossmodal AI Framework for Smart Data Analysis}
\label{MCAI}

We have researched and developed the Multimodal and Crossmodal AI Framework (MMCRAI) to contribute to the evolution of multimodal and cross-modal AI in smart data analysis. This framework's significant advantage is creating a hybrid backbone that can flexibly be re-constructed in different ways to build a suitable individual model for a particular problem. The framework is designed to take into account the following criteria:
\begin{itemize}
    \item Strengthen the robust prediction by enhancing similar modalities (i.e., multi-sensors capture the same data)
    \item Enhance robust inferences and generate new insights from different modalities by carrying complementary information about each other during the learning process (e.g., fusion, alignment, co-learning). 
    \item Establish cross-modal inferences \cite{crossmodal_ravela2005} to overcome noisy and missing data of one modality by using information (e.g., data structure, correlation, attention) found in another modality.
    \item Discover cross-modal attention to enhance cross-modal search (e.g., language-video retrieval, image captioning, translation) or ensure the semantic-harmony among modalities (e.g., cheapfakes detection).
\end{itemize}

Currently, multimodal and crossmodal approaches work independently due to the domain-dependently intentional-architecture design. Hence, a framework that allows people to integrate multimodal and crossmodal into a uni-progress can enhance the scaling and the ability to connect or incorporate different uni-progresses to solve multi-domain problems. 

\begin{figure}[hbt!]
    \centering
    \includegraphics[width=\textwidth]{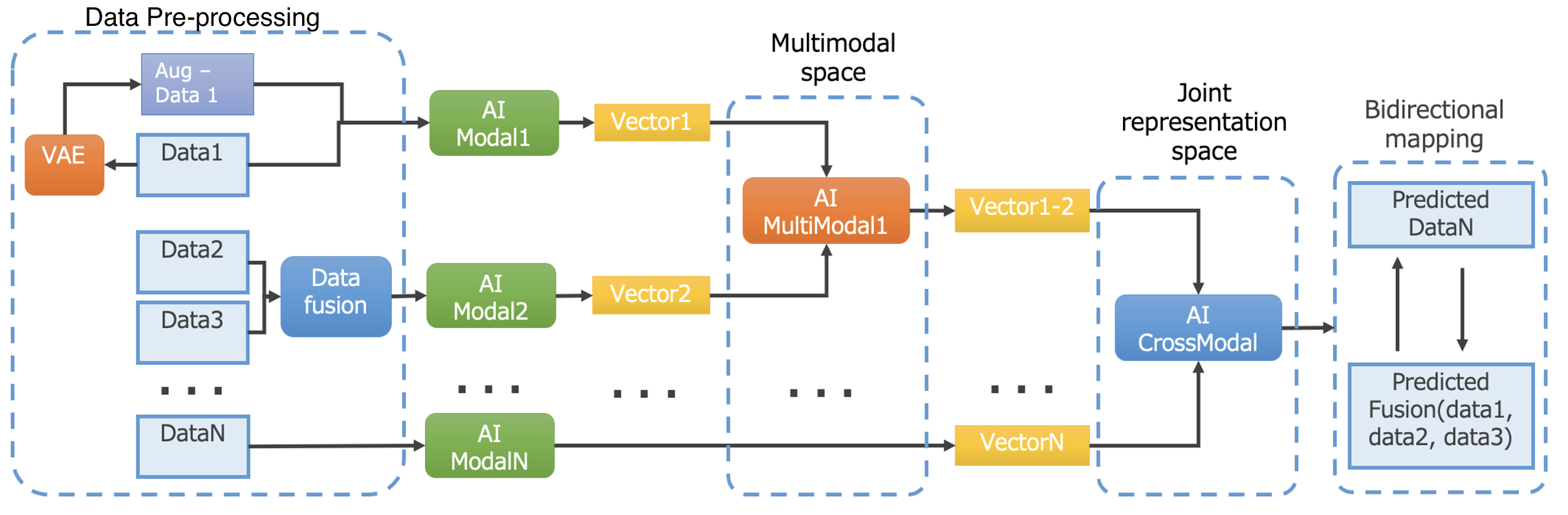}
    \caption{A general design of the multimodal and crossmodal AI framework MMCRAI (VAE: Variational Autoencoders, Aug-Data: Augmented Data)}
    \label{fig:img_1}
\end{figure}

We design the framework as the hierarchical structure of multimodal and crossmodal approaches where a suitable approach can be utilized depending on the purpose of application. Figure \ref{fig:img_1} illustrates the framework's general structure that aims to create a joint multimodal representation by embedding every single-modal representation into a common representation space. The design starts with the data pre-processing component. Here disjoint modalities are gathered and pre-proceed, such as cleansing, fusing, and augmenting. Next, those modalities that do not need to have a crossmodal translation (i.e., bidirectional mapping) are sent to the multimodal space component. Applications requiring retrieval and classification tasks without translating from the multimodal representation to the single-modal ones can utilize this component without going further. Applications that require crossmodal translation in addition to classification tasks, such as multimodal query expansion and crossmodal retrieval, should go to the joint representation space and bidirectional mapping components.  

Based on this general structure, we have developed an MM-sensing family with two representatives, MM-AQI and MM-trafficEvent, and 3DCNN for dealing with air pollution, safety driving, and congestion problems. While 3DCNN and MM-trafficEvent focus on multimodal and crossmodal approaches, MM-AQI mixes both techniques. 

\section{MM-sensing}
\label{MMS}

The MM-sensing stands for Multimedia Sensing, a virtual intelligent sensor that can predict complex events in the real world from multimodal observation data such as images, videos, sensory data, and texts. As represented in the name, MM-sensing mainly deals with multimedia data that occupy a significant portion of data due to the explosion of multimedia IoT devices and the high-speed bandwidth of the Internet (e.g., 5G, 6G). Another reason to build MM-sensing is that multimedia data contain vast semantic meaning that is hard to extract. Hence, it could be good to have an independent component that can provide high-semantic information to the other processes or applications. 

The following subsections will explain how to downstream the general framework into different applications running in various domains. We introduce MM-AQI and MM-trafficEvent as two downstream versions of the general framework working in air pollution prediction and traffic incident querying.

\subsection{MM-AQI: a crossmodal AI for estimating air quality index from lifelog images}


Air pollution harmfully impacts human life, including health, economy, urban management, and climate change \cite{aqi_lu2020}. nfortunately, air pollution prediction is not a trivial problem that can predict a new value using a sole data source. Many factors can impact the air pollution prediction, such as human activities (e.g., transportation, mining, construction, industrial and agricultural activities), weather (e.g., winds, temperature, humidity), and natural disasters (e.g., volcano eruptions, earthquakes, wildfires). Unfortunately, using data captured by individual modalities may not gather complementary information to express the correlation and causality among factors with air pollution. Hence, the approach of multimodal learning that consolidates multi modalities from various mentioned factors into a single air pollution prediction model has become popular \cite{aqi_dat2021}\cite{aqi_zhao2020}\cite{aqi_liang2020}.  

Although many methods have been established to monitor and predict air pollution, an eco-friendly and personal-usage method is still the most significant challenge. Expensive and large-scale deployed devices that provide high-quality air pollution data do not exist in a dense grid in developed countries, and the situation can be worse in developing and emerging countries. Besides, to cope with such big multimodal data, there is a need for supercomputers or luxury GPU servers, which makes it hard to have an eco-friendly and personal-usage application for personal usage. 

To cope with this challenge, we design a crossmodal AI, MM-AQI, a member of the MM-sensing family that can estimate the current air pollution level (i.e., PM2.5) using lifelog images. Lifelog images are a set of images captured periodically for a long time by a personal camera. MM-AQI hypothesizes that lifelog images may contain information that correlates to air pollution. Hence, we can use only lifelog images, plus position and time data, to predict air pollution. The scenario is that a user takes a picture with his/her smartphone, or a personal camera that can connect to the smartphone, the crossmodal AI installed in the smartphone will estimate the current air pollution level. First, we use multimodal datasets collected from air pollution stations, mobile devices, and lifelog cameras over a particular area to analyze the correlation between the surrounding environment (e.g., human activities, weather, natural disasters) and air pollution. This step helps us decide which features extracted from images can correlate with the air quality index. In other words, we discover the cross attention between image features and air quality index values. Then, we build a crossmodal AI to predict the air quality index using only lifelog images. That meant, we need crossmodal translation to translate implicit human activities, weather, and natural disasters from lifelogging images to air quality index values.

\begin{figure}[hbt!]
    \centering
    \includegraphics[width=0.8\textwidth]{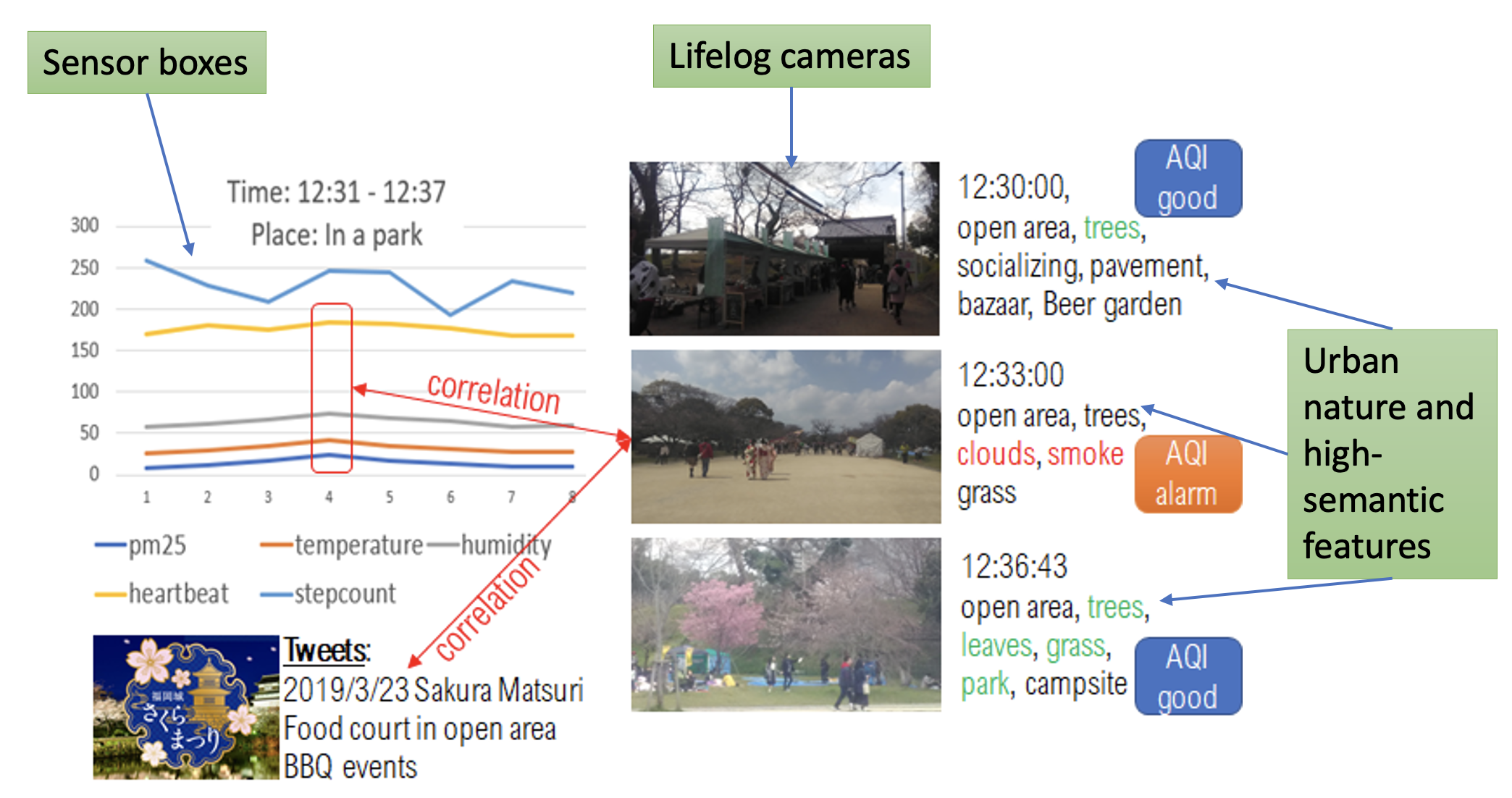}
    \caption{The correlation between the surrounding environment (e.g., human activities, natural disasters) and air pollution \cite{aqi_dao2021}}
    \label{fig:img_2}
\end{figure}

\begin{figure}[hbt!]
    \centering
    \includegraphics[width=0.8\textwidth]{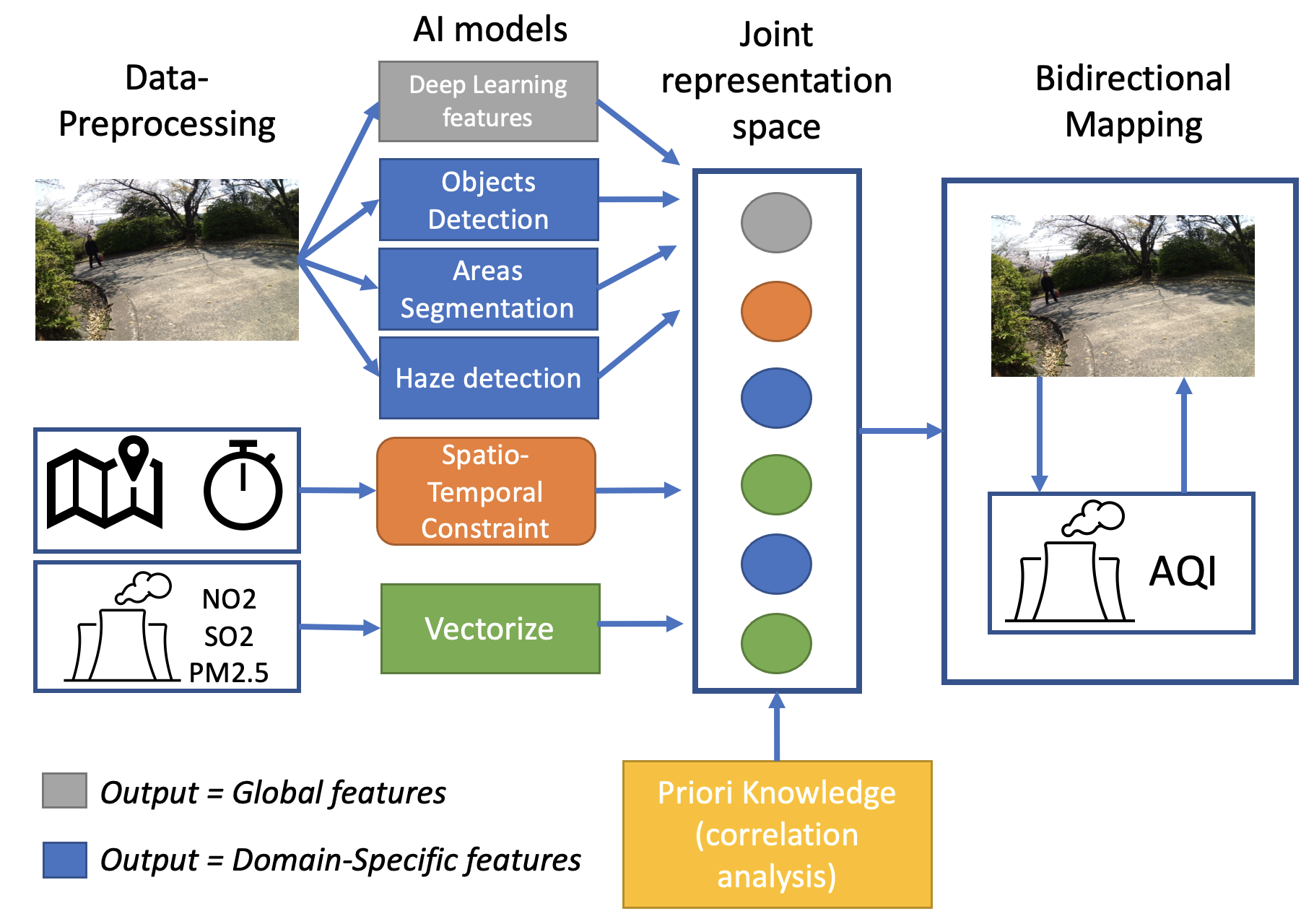}
    \caption{MM-AQI: The application-wise design}
    \label{fig:img_2_1}
\end{figure}

Figure \ref{fig:img_2} illustrates the results of correlation analysis between the surrounding environment and air pollution. We recognize that lifelog images and air pollution captured and measured at the same spatiotemporal dimension can map from one to another and back and represent them in joint representation space. Hence, we modify the MM-sensing general structure to create the MM-AQI crossmodal AI to predict air pollution using lifelog images, as depicted in Figure \ref{fig:img_2_1}. We keep the data-preprocessing, joint representation space, and bidirectional mapping components. In contrast, the multimodal space component is replaced by the prior knowledge created by learning the correlation among images and AQI at the same place over time. Based on the correlation analysis on a historical dataset of both lifelog images and AQI, we can build the prior knowledge to help us design global and domain-specific features. As shown in Figure \ref{fig:img_2}, the area features such as the green zone, dirty zone, street zone, and sidewalk zone, and object features such as vehicles, trees, pedestrians, and haze have a tight correlation with the fluctuation of AQI value. Hence, we take these features as domain-specific while using deep learning embedding vectors as global to consider both low and high-semantic features in the joint representation space. The key idea here is to estimate complete AQI from noisy and missing observations of one sensory modality (i.e., air pollution device) using a structure found in another (i.e., images). In other words, with this model, we can estimate the AQI value by simply using all features extracted from an image mapped and linked to a proper AQI level inside the joint representation space and by the bidirectional mapping. 

\begin{figure}[hbt!]
    \centering
   \includegraphics[width=0.7\textwidth]{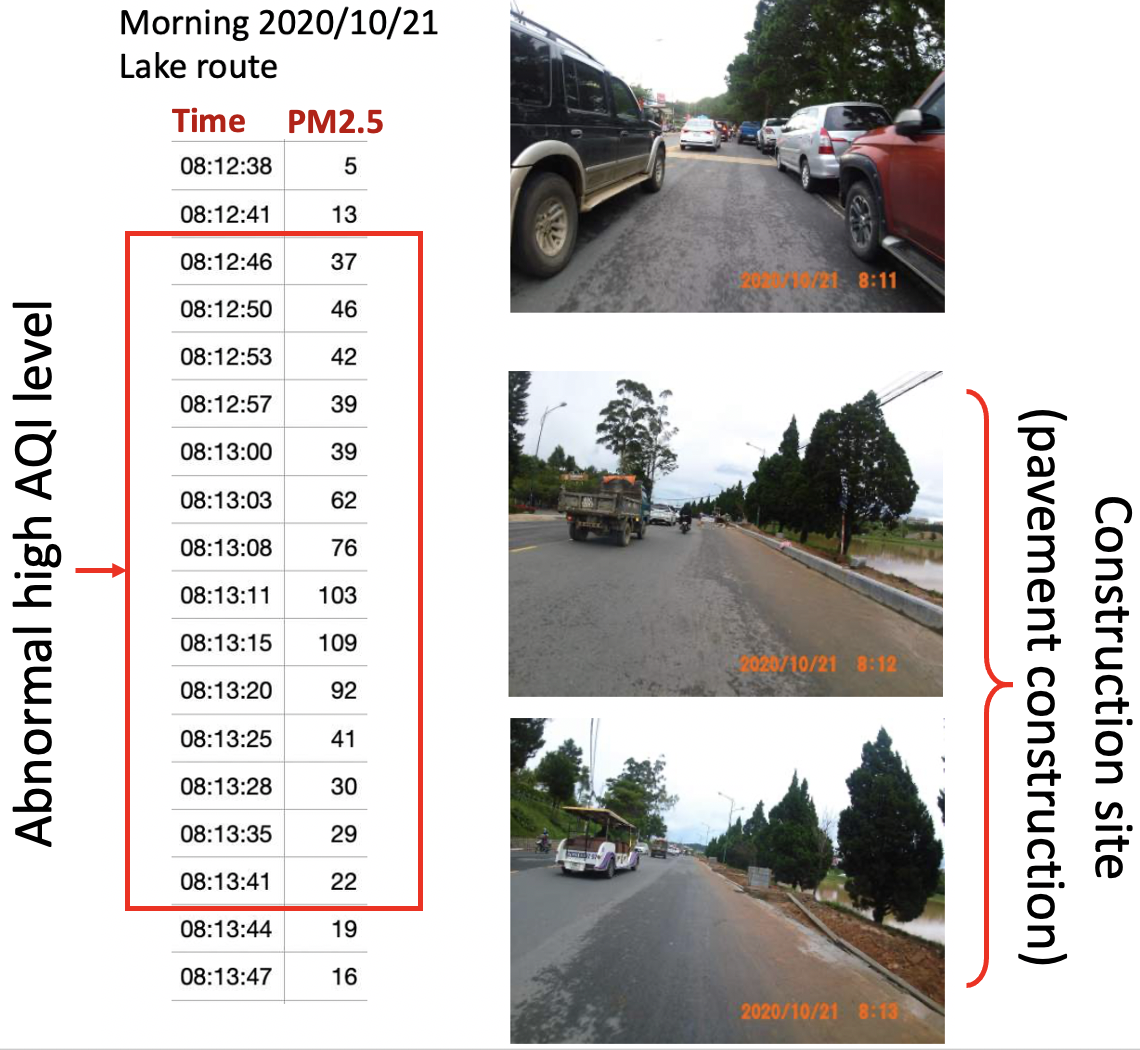}
    \caption{MM-AQI: An example of high PM2.5 and human activities captured by lifelog camera\cite{aqi_dao2021}. In this case, objects =  \{vehicles \}, areas =  \{dirty/dust\}, and abnormal high value of PM2.5 appear at the same place and time.}
    \label{fig:img_cor}
   \vspace{-4mm}
\end{figure}

The evaluation conducted using three different datasets collected from Japan, Vietnam, and India provides an accuracy of over $80\%$ (F1-score). That is an impressive result when running on a low-cost device (e.g., smartphones). Besides, MM-AQI not only estimates the air quality index but also offers several cues to understand the causality of air pollution (Figure \ref{fig:img_cor}. Thanks to MM-sensing flexible architecture, both autoencoder-decoder and transformer architectures can be applied to design MM-AQI crossmodal. For more details, readers can refer to the original paper of MM-AQI \cite{aqi_dao2021} and its extension version \cite{aqi_vinh2021}.

\begin{table}[hbt!]
\begin{center}
\caption{PM2.5 Prediction Accuracy Comparison (F1-score)}
\begin{tabular}{|c|c|c|c|} \hline
\textbf{Dataset} & \textbf{Our Model} &  \textbf{LSTM-based Prediction\cite{aqi_dao2021}} \\
\hline
MNR-HCM & 0.88 &  0.73  \\
VisionAir India &  0.87 & 0.82\\
Japan & 0.92 & 0.84\\ \hline
\end{tabular}
\label{tab:eval2}
\end{center}
\end{table}

\subsection{MM-trafficEvent: A crossmodal with attention to query images from textual queries}

\begin{figure}[hbt!]
    \centering
    \includegraphics[width=\textwidth]{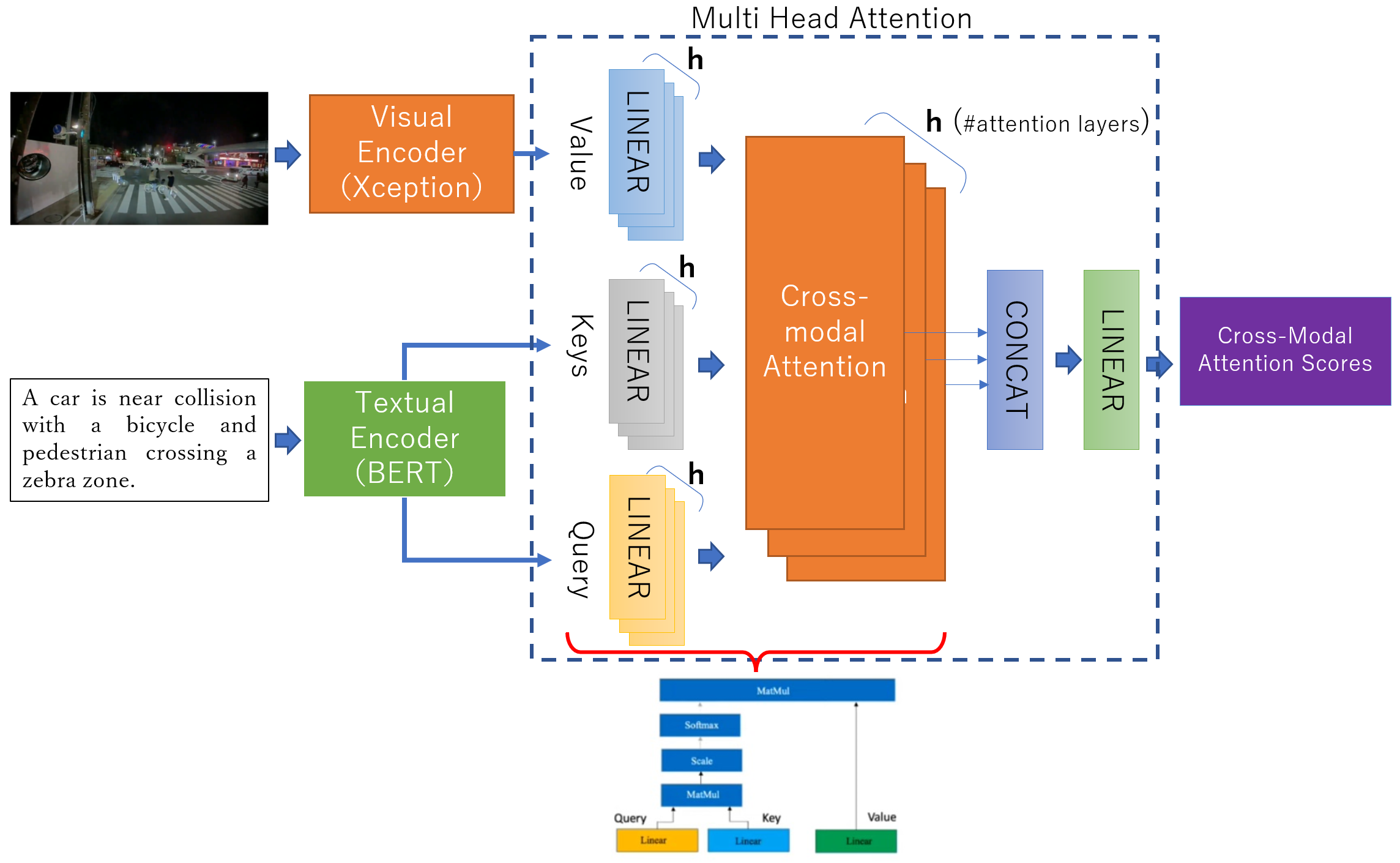}
    \caption{MM-trafficEvent: A cross-modal multi-head attention model}
    \label{fig:img_5}
\end{figure}

Dashcam, a video camera mounted on a vehicle, has become a popular and economical device for increasing road safety levels \cite{traffic_Adamova2020}. A new generation of the dashcam, the smart dashcam, not only records all events happening during a journey but also alerts users (e.g., drivers, managers, coaches) of potential risks (e.g., crash, near-crash) and driving behaviors (e.g., distraction, drowsiness). One of the significant benefits of dashcam footage is that it can provide insights from dashcam data to support safe driving \cite{traffic_kim2020}(e.g., evidence to the police and insurance companies in traffic accidents, self-coaching, fleet management). Unfortunately, a significant obstacle to events retrieval is a lack of searching tools for finding the right events from a large-scale dashcam database. The conventional approach to finding an event from dashcam footage is to manually browse a whole video from beginning to end. It consumes a lot of workforces, time, and money. The challenge is the semantic gap between textual queries made by users and visual dashcam data. It needs a crossmodal translation to enable the ability to retrieve related data of one modality (e.g., dashcam video shots) with data of another modality (e.g., textual queries) \cite{crossmodal_zhang2022}\cite{crossmodal_Ji202}. 

To provide a user-friendly tool that can support users in quickly finding an event they need, we introduce MM-trafficEvent as a text-image crossmodal with an attention search engine by modifying the MMCRAI general architecture. Figure  \ref{fig:img_5} illustrates the design of this function. First, we replace "AI modal" modules with encoder models (i.e., Xception for image, BERT for text) that aim to normalize and polish raw data into vector spaces. Second, we design the attention mechanisms as the joint representative space to provide an additional focus on a specific area that has the same mapping from different modalities. In other words, we utilize self and multi-head attention techniques \cite{crossmodal_Vaswani2017} to generate the bidirectional mapping between text and image. In this design, we replace the joint representative space and bidirectional mapping with multi-head attention block and cross-modal attention scores, depicted in Figure \ref{fig:img_5}. 

The significant difference between our model and others is that we do not have a full training dataset of text-image pairs. In other words, we do not have the annotation/caption of each incident/suspect event image. Hence, creating a complete text-image crossmodal retrieval is almost impossible by using only the dashcam video dataset. Instead of applying crossmodal translation directly, we utilize word-visual attention pre-trained model for transforming the query and the dataset to the same space (i.e., joint representative space). We use available text-image pairs open datasets with our datasets collected during interactive querying with users for downstream the pre-trained model to adapt to our domain. Hence, when users input their textual queries, we utilize our crossmodal text-image to find a set of sample images used again as visual queries to search over the dashcam dataset.

We use various datasets gathered from public sources and created by ourselves. We asked two volunteers to label data and created a structured data set $D_k = \{(I_k; [T^{i}_{k}])\}$ where $I_k$ denotes $i^th$ image, and $[T^{i}_{k}]$ represents a list of captions similar with that image. In our dataset, we have only a set of $D_k = \{(I_k; [T_{k}])\}$ and in practice, we also use a dataset from RetroTruck and I4W datasets with a set of 5 captions for each image to generate the pre-trained weight set. 


\begin{figure}[hbt!]
    \centering
    \includegraphics[width=\textwidth]{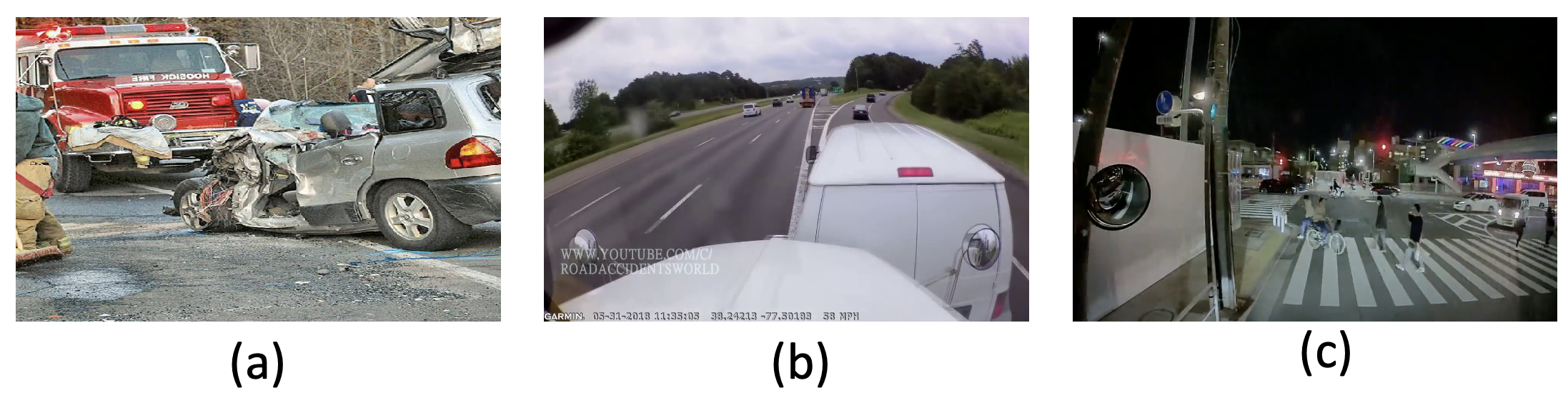}
    \caption{MM-trafficEvent: A sample of query-result outputs (a) Q: "find an accident made by a van and a red bus" [semantic level = easy], (b) Q: "find an accident where a white truck hit a white van from behind" [semantic level = complex], (c) Q: "find a moment a truck stop closed to the zebra zone where a lot of pedestrians and bicycles are crossing" [semantic level = most complex]. (a)(b) from I4W and  RetroTruck datasets, (c) from our dataset}
    \label{fig:img_fine}
\end{figure}

Table \ref{tab:Free-style-Query} shows that the system has good productivity when users almost found their results in the first round with the average of P@10 as 7.18 (i.e., seven relevant results over ten retrieved results at the first try). Statistically, the system works well when finding the expected results within 15 loops with naive users and 10 loops with expert users. Besides, the simulation results confirm the interactive GUI’s an advantage when decreasing the P@K from 200 to 10. Figure \ref{fig:img_fine} illustrates one example of events retrieved by our system with different difficult levels of semantic levels of textual queries (i.e., easy, complex, most complex). For more details, readers can refer to the original paper of MM-trafficEvent \cite{traffic_nguyen2020}\cite{traffic_dao2021}

\begin{table}[htb]
    \centering
    \caption{Incident querying results using MM-trafficEvent model \cite{traffic_dao2021}}
    \label{tab:Free-style-Query}
    \begin{tabular}{ll}
        \hline
        \textbf{Parameter} &  \textbf{Value}\\ 
        \hline
        RetroTruck and I4W datasets & 854 videos ($1\sim15$ mins/each)\\
        Number of Queries & 50\\
        Evaluation Metric & P@10  \\
        Average result by naive user & Avg(P@10) = 7.18/10 $\approx 72\%$\\
        Average result by expert user & Avg(P@10) = 7.56/10 $\approx 76\%$\\
        Average finding-loops by naive user & 15.3\\
        Average finding-loops by expert user & 10.8\\
        \hline
    \end{tabular}
\end{table} 

\section{3DCNN: a multimodal AI model for spatio-temporal event prediction}




The typical approach to dealing with multimodal data is data fusion, which aims to collect significant fragments of an object distributed in different modalities and normalize these fragments into the same space for easier manipulation. Three fusion methods are popular: early, late, and collaborative fusion\cite{multimodal_wang2018}. The first one fuses data first and processes the fuse data with specific models. The second one performs data analysis of each information independently with a particular model, then combines the outputs as the outcome. The last one aims to promote collaboration among modalities to achieve the ideal consensus. Hence, we introduce a 3DCNN model with a specific collaborative fusion mechanism, called raster-images, to fuse spatio-temporal multimodal data into one unique data format that up-to-date computer vision deep learning models can efficiently utilize.  

Unlike other multimodal methods working with spatio-temporal data, we want to embed the spatio-temporal dimension into our model without converting them into an alternative space. In other words, we want to keep the geometry topology (i.e., time and location) and map other data into this coordination before projecting a whole dataset to other spaces. To do that, we convert the MMCRAI framework to have two main components: spatio-temporal-based data wrapping and multimodal space working as collaborative fusion and multimodal embedding.  

\begin{figure}[hbt!]
    \centering
    \includegraphics[width=0.8\textwidth]{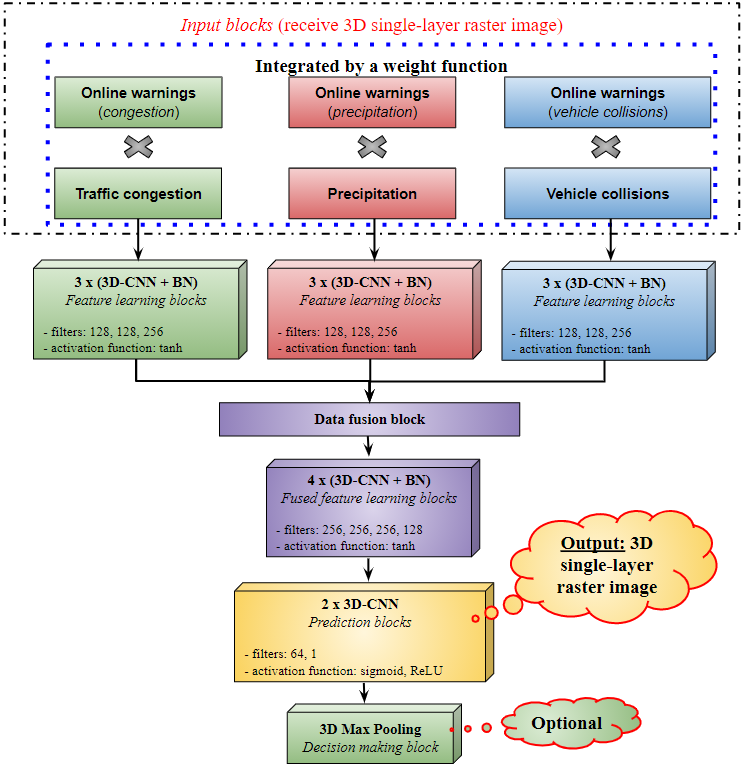}
    \caption{ 3D-CNN Multi-sources data deep learning architecture \cite{3d_dao2021}}
    \label{fig:img_3}
\end{figure}

To conduct the collaborative fusion, we invent a new fusion schema that can wrap different modalities into one unique spatial modality, namely a raster image. Using hashing techniques, we distribute multimodal data collected within a specific time window into three channels (R, G, B) of a raster image whose pixels represent particular map areas (i.e., spatio-temporal constraints). When arranging these raster images along a time dimension, we produce a so-called raster video whose frames are raster images. Hence, the raster image perfectly replaces the multimodal space component of the general framework and guarantees the consensus of modalities projected on the same spatio-temporal dimension. At this stage, we can apply AI models (e.g., CNN, LSTM, RNN, Transformer) to create multimodal embedding that will be utilized to predict events. In our case, we develop a model based on 3D convolutional neural networks (3DCNN) architecture that can extract necessary features from a raster video input to form the multimodal embedding. Figure \ref{fig:img_3} illustrates one of our 3DCNN models published in IEEE Big Data 2019 \cite{3d_dao2019}. 


To demonstrate the efficiency and influence of the 3DCNN, we apply this model to predict congestion using open data sources that can easily access in cyberspace. We leverage our observation of the correlation between bad weather (e.g., heavy rain, flood, snow), traffic congestion, and human behavior (e.g., claim about congestion and/or bad weather on SNS) during the bad period to decide which data sources to be imported to our model. Hence, we utilize data captured from social networks (e.g., Twitter), meteorology agencies (e.g., XRAIN precipitation data www.diasjp.net), and traffic agencies (e.g., Traffic Congestion Statistics Data www.jartic.or.jp) as our multi modalities. 

The reason for choosing the congestion prediction topic is that congestion is one of the most prevalent transport challenges in large urban agglomerations \cite{3d_Rodrigue2020}. Therefore, a robust prediction of congestion and congestion-surrounding-environment correlation discovery using data from different sources became the most significant demand from society. Many researchers have developed several multimodal AI models to predict congestion \cite{3d_akhtar2021}\cite{3d_kumar2021}\cite {3d_jiang2021}. The common idea of these methods is to create a joint multimodal representation by embedding every single-modal representation into a common representation space, with or without constraints of time and locations. Compared to these methods, the 3DCNN model significantly differs by turning the multimodal space or joint representation space into a popular data format (i.e., videos) and the ability to wrap unlimited modalities into one space without any extra activities. 


\begin{figure}[hbt!]
    \centering
    \includegraphics[width=\textwidth]{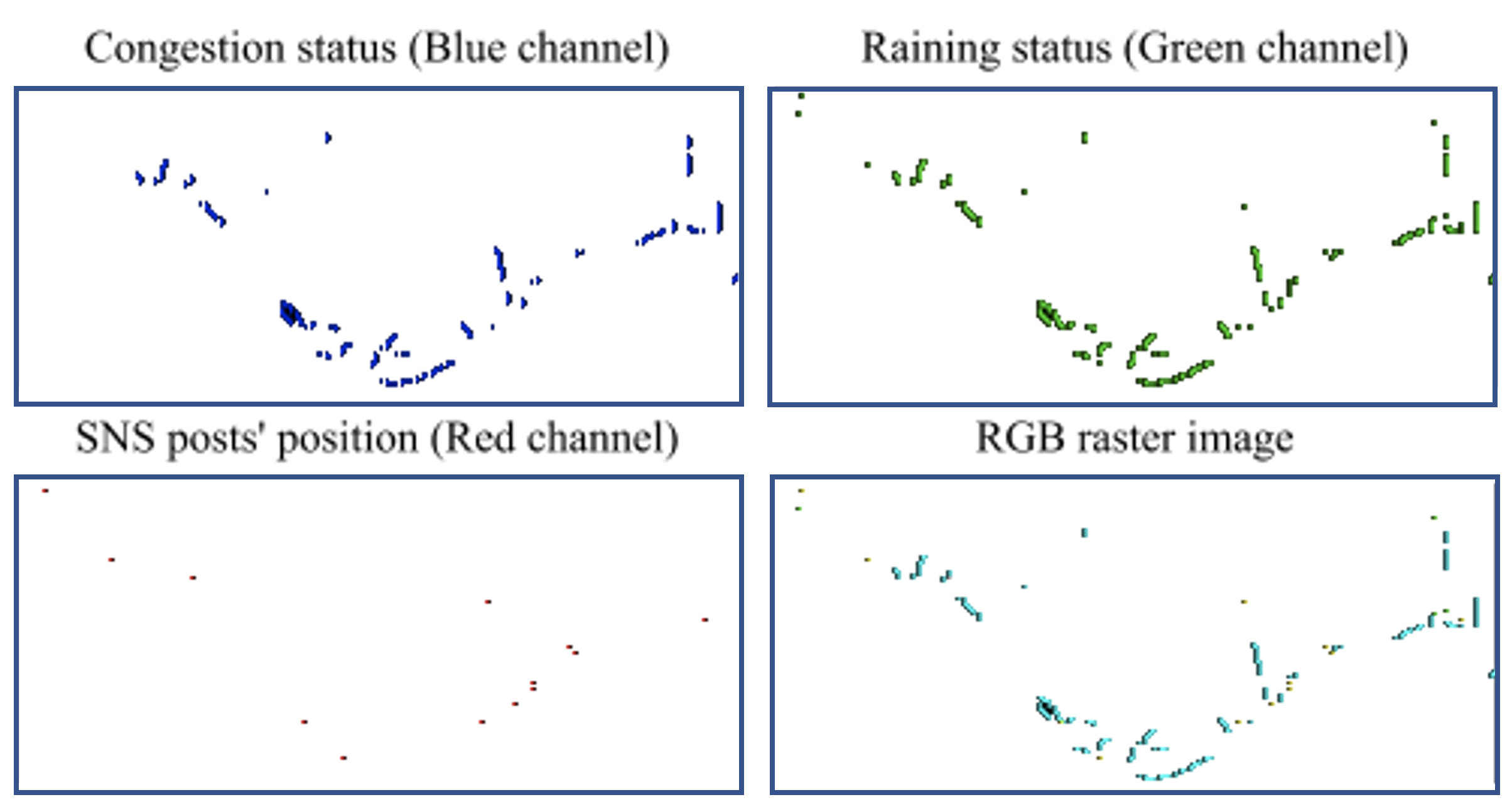}
    \caption{Collaborative Fusion by wrapping data into a raster video \cite{3d_dao2019}}
    \label{fig:img_raster}
\end{figure}

Figure \ref{fig:img_raster} depicts how to wrap three different modalities into one space. First, we convert each modality into an individual channel by picking a data value from one mesh code and convert it into one pixel. The (longitude, latitude) of the mesh code is mapped into an image's (width, height), and the data value is normalized into [0, 255].  Then, we merge these channels to make the raster image (R, G, B).

\begin{figure}[hbt!]
    \centering
    \includegraphics[width=\textwidth]{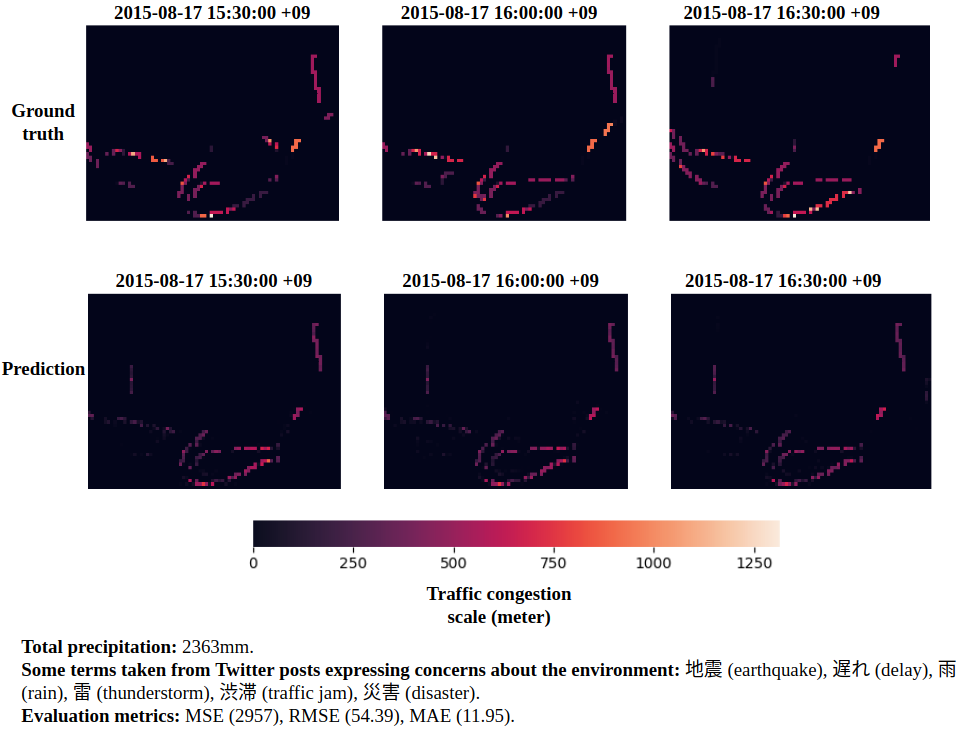}
    \caption{Short-term predicting results using 3D-CNN model on traffic congestion, precipitation, and tweets data \cite{3d_dao2019}, based on the raster image created as in Figure \ref{fig:img_raster}}
    \label{fig:img_4}
\end{figure}

\begin{table}[htb!]
\caption{Traffic congestion prediction models comparison\cite{3d_dao2021} (measured by MAE, lower is better)}
\begin{center}
\begin{tabular}{|c|c||c|c|}
 \hline 
 \textbf{Model} & \textbf{MAE} & \textbf{Model} & \textbf{MAE} \\
 \hline\hline
 Historical Average & 10.24 & Seq2Seq AT+NB \cite{Liao_2018} & 8.75\\
 \hline
 Vector Autoregression & 9.44 &  Fusion-3DCNN (this work) & \underline{\textbf{8.13}} \\
 \hline
 \end{tabular}
\label{table_fusion3dcnn}
\end{center}
\end{table}

Figure \ref{fig:img_4} depicts one example of using 3DCNN to predict short-term congestion using traffic congestion, weather, and tweets data over the Kobe-Japan area. As depicted in Figure \ref{fig:img_4}, each picture is a single-channel raster image instance containing only congestion information over the transportation network of the Kobe area. Each pixel of this image reports the congestion level of one mesh code. The brighter color is the heavier congestion. The latest version of 3DCNN gains outstanding results with MAE=8.13 compared to other methods, as denoted in Table \ref{table_fusion3dcnn}. For more details, readers can refer to the original paper of 3DCNN \cite{3d_dao2019} and its last extension version \cite{3d_dao2021}.

\section{Conclusions}
\label{CON}

In this paper, we comprehensively discuss the vital role of multiple modalities of data and crossmodal AI techniques in understanding the surrounding world. We also introduce our multimodal and crossmodal AI framework for smart data analytics. Moreover, we present three instances of this framework to tackle air pollution, traffic incident query, and congestion prediction problems. For each instance, we discuss the motivation and hypothesis by which we can adjust the general framework to adapt different multimodal datasets and crossmodal AI to solve the problem. In the future, we will continue to extend the framework and develop accurate AI models to deal with more challenges. We also want to investigate more on making the framework able to work in a mobile environment (e.g., IoT devices) and distributed networks (e.g., Federated Learning). 
%
%


\begin{thebibliography}{6}
%
\bibitem {multimodal_Baltrusaitis2019}
Baltrusaitis, T., Ahuja, C., Morency, L.P.: Multimodal Machine Learning: A Survey and Taxonomy. IEEE Trans. Pattern Anal. Mach. Intell. 41, 2 (February 2019), 423–443.

\bibitem {multimodal_bayoudh2022}
Bayoudh, K., Knani, R., Hamdaoui, F. et al: A survey on deep multimodal learning for computer vision: advances, trends, applications, and datasets. Vision Computing, volume 38, 2939–2970 (2022).

\bibitem {modal_Vukotic2016}
Vukotić, V., Raymond, C., Gravier, G.: Bidirectional Joint Representation Learning with Symmetrical Deep Neural Networks for Multimodal and Crossmodal Applications. In Int. Conf.e on Multimedia Retrieval (ICMR '16). 

\bibitem {crossmodal_wang2016}
Wang, K.Y., Yin, Q.Y., Wang, W., Wu, S., Wang, L: A Comprehensive Survey on Cross-modal Retrieval. CoRR abs/1607.06215 (2016)

\bibitem {crossmodal_Ji202}
Ji et. al: CRET: Cross-Modal Retrieval Transformer for Efficient Text-Video Retrieval. SIGIR ’22, July 11–15, 2022, pp. 949-959. 

\bibitem {crossmodal_Khare2021}
Khare, A., Parthasarathy, S., Sundaram, S.: Self-Supervised Learning with Cross-Modal Transformers for Emotion Recognition. In IEEE Spoken Language Technology Workshop (SLT), 2021, pp. 381-388

\bibitem {crossmodal_zhang2022}
Zhang, J.W., Wermter, S., Sun, F.C., Zhang, C.S., Engel, A.K, Röder, B., Fu, X.L.:
Editorial: Cross-Modal Learning:
Adaptivity, Prediction and Interaction. Frontiers in Neurorobotics, Volume 16, Article 889911, April 2022.

\bibitem {crossmodal_ravela2005}
Ravela, S., Torralba, A., Freeman,  W. T.: An ensemble prior of image structure for cross-modal inference, Tenth IEEE International Conference on Computer Vision (ICCV'05) Volume 1, 2005, pp. 871-876 Vol. 1

\bibitem {aqi_lu2020}
Lu, G.J: Air pollution: A systematic review of its psychological, economic, and social effects. Current Opinion in Psychology, Volume 32, 2020, Pages 52-65.

\bibitem {aqi_dat2021}
Duong, Q.D., Le, M.Q., Nguyen-Tai, T.L., Nguyen, D.H., Dao, M.S., Nguyen, T.B.:
An Effective AQI Estimation Using Sensor Data and Stacking Mechanism. SoMeT 2021: 405-418

\bibitem {aqi_zhao2020}
Zhao, P., Zettsu,K.: MASTGN: Multi-Attention Spatio-Temporal Graph Networks for Air Pollution Prediction. IEEE Big Data 2020, pp.1442-1448, 2020.

\bibitem {aqi_liang2020}
Liang, Y.C., Maimury, Y., Chen, A.L., Juarez J.R.C.: Machine Learning-Based Prediction of Air Quality. Applied Sciences. 2020, 10(24):9151

\bibitem {aqi_dao2021}
Dao, M.S., Zettsu, K., Uday, R.K.:
IMAGE-2-AQI: Aware of the Surrounding Air Qualification by a Few Images. IEA/AIE (2) 2021: 335-346

\bibitem {aqi_vinh2021}
La, T.V., Dao, M.S., Tejima, K., Uday R.K., Zettsu, Z.:
Improving the Awareness of Sustainable Smart Cities by Analyzing Lifelog Images and IoT Air Pollution Data. IEEE BigData 2021: 3589-3594

\bibitem {traffic_Adamova2020}
Adamová, V.: Dashcam as a device to increase the road safety level. In Int. Conf. on Innovations In Science and Education (CBU), 2020, pp. 1–5

\bibitem {traffic_kim2020}
Kim, J., Park, S., Lee, U.: Dashcam witness: Video sharing motives and privacy concerns across different nations. IEEE Access, vol. 8, pp. 425–437, 2020.

\bibitem {crossmodal_Vaswani2017}
Ashish Vaswani, Noam Shazeer, Niki Parmar, Jakob Uszkoreit, Llion Jones, Aidan N. Gomez, Łukasz Kaiser, and Illia Polosukhin: Attention is all you need. In Proceedings of the 31st International Conference on Neural Information Processing Systems (NIPS'17). 

\bibitem {traffic_nguyen2020}
Mai-Nguyen, A.V., Phan, T.D., Vo, A.K., Tran, V.L., Dao, M.S., Zettsu, K.:
BIDAL-HCMUS@LSC2020: An Interactive Multimodal Lifelog Retrieval with Query-to-Sample Attention-based Search Engine. LSC@ICMR 2020: 43-49

\bibitem {traffic_dao2021}
Dao, M.S., Pham, D.D., Nguyen, M.P., Nguyen, T.B., Zettsu, K.:
MM-trafficEvent: An Interactive Incident Retrieval System for First-view Travel-log Data. IEEE BigData 2021: 4842-4851


\bibitem {multimodal_wang2018}
Wang, Y.: Survey on Deep Multi-modal Data Analytics: Collaboration, Rivalry and Fusion. J. ACM 37, 4, Article 111 (August 2018).

\bibitem {3d_dao2019}
Dao, M.S., Nguyen, N.T., Zettsu, K.: Multi-time-horizon Traffic Risk Prediction using Spatio-Temporal Urban Sensing Data Fusion. 2019 IEEE International Conference on Big Data (Big Data).

\bibitem {3d_Rodrigue2020}
Rodrigue, J.-P.: The Geography of Transport Systems, FIFTH EDITION, New York: Routledge, 456 pages, 2020. ISBN 978-0-367-36463-2

\bibitem {3d_akhtar2021}
Akhtar, M., Moridpour, S.: A Review of Traffic Congestion Prediction Using Artificial Intelligence. Journal of Advanced Transportation, vol. 2021, Article ID 8878011, 18 pages, 2021.

\bibitem {3d_kumar2021}
Kumar, N., Raubal, M.: Applications of deep learning in congestion detection, prediction and alleviation: A survey. Transportation Research Part C: Emerging Technologies,
Volume 133, 2021, 103432.

\bibitem {3d_jiang2021}
Jiang, H.L., Li, Q., Jiang, Y., Shen, G.B., Sinnott, R., Tian, C., Xu, M.G.: When machine learning meets congestion control: A survey and comparison. Computer Networks, Volume 192, 2021, 108033, ISSN 1389-1286.


\bibitem {3d_dao2021} 
Dao, MS., Uday Kiran, R., Zettsu, K.: Insights for Urban Road Safety: A New Fusion-3DCNN-PFP Model to Anticipate Future Congestion from Urban Sensing Data. In: Kiran, R.U., Fournier-Viger, P., Luna, J.M., Lin, J.CW., Mondal, A. (eds) Periodic Pattern Mining . Springer

\end{thebibliography}
\end{document}